\documentclass[conference]{IEEEtran}
\IEEEoverridecommandlockouts

\usepackage{cite}
\usepackage{amsmath,amssymb,amsfonts}
\usepackage{algorithmic}
\usepackage{graphicx}
\usepackage{textcomp}
\usepackage{xcolor}
\usepackage{multirow}
\usepackage{threeparttable}
\usepackage{booktabs}
\usepackage{epstopdf}
\usepackage[T1]{fontenc}
\usepackage{aecompl}
\usepackage{hyperref}
\usepackage{cleveref}

\def\BibTeX{{\rm B\kern-.05em{\sc i\kern-.025em b}\kern-.08em
    T\kern-.1667em\lower.7ex\hbox{E}\kern-.125emX}}
\begin{document}

\title{Lithium-ion Battery State of Health Estimation \protect\\based on Cycle Synchronization \protect\\ using Dynamic Time Warping\\
}

\author{Kate Qi~Zhou,~\IEEEmembership{}
	Yan~Qin,~\IEEEmembership{Member,~IEEE,}
	Billy Pik Lik~Lau,~\IEEEmembership{}
	Chau~Yuen,~\IEEEmembership{Fellow,~IEEE,}
         Stefan~Adams
	\thanks{This work was supported by EMA-EP011-SLEP. (Corresponding author: Qin Yan)}
	\thanks{KQ. Zhou, Y. Qin, BPL.Lau and C. Yuen are with the Engineering Product Development Pillar, The Singapore University of Technology and Design, 8 Somapah Road, 487372 Singapore. (e-mail: qi$\_$zhou@mymail.sutd.edu.sg, yan.qin@ntu.edu.sg, yuenchau@sutd.edu.sg)}
	\thanks{S.Adams is with the Materials Science and Engineering, National University of Singapore, 117575 Singapore. (e-mail: mseasn@nus.edu.sg)}}


\maketitle
\begin{abstract}
The state of health (SOH) estimation plays an essential role in battery-powered applications to avoid unexpected breakdowns due to battery capacity fading. However, few studies have paid attention to the problem of uneven length of degrading cycles, simply employing manual operation or leaving to the automatic processing mechanism of advanced machine learning models, like long short-term memory (LSTM). As a result, this causes information loss and caps the full capability of the data-driven SOH estimation models. To address this challenge, this paper proposes an innovative cycle synchronization way to change the existing coordinate system using dynamic time warping (DTW), not only enabling the equal length inputs of the estimation model but also preserving all information. By exploiting the time information of the time series, the proposed method embeds the time index and the original measurements into a novel indicator to reflect the battery degradation status, which could have the same length over cycles. Adopting the LSTM as the basic estimation model, the cycle-synchronization-based SOH model could significantly improve the prediction accuracy by more than 30$\%$ compared to the traditional LSTM.
\end{abstract}


\section{Introduction}
Lithium-ion batteries (LiB) have been a crucial component in many applications like home appliances, electric vehicles, and industrial factories [1]. Technological breakthroughs have enabled the development of cost-effective, long-lasting LiB compared to decades ago. However, LiB performance degrades over cycles through usage [2]. Therefore, state of health (SOH), which deals with the battery prognostics to reach a failure level [3], has gained massive attention in both the academic and industrial space.

Battery degradation belongs to cyclic degradation, which follows an inconsistently repetitive but similar degradation behavior for the same object from cycle to cycle [4]. Capacity is commonly used as an indicator to indicate degradation behavior [5]. Usually, there are two groups of approaches on LiB capacity estimation, which are the model-based methods and data-driven methods. Model-based methods require much of prior physical or electrochemical knowledge to model degradation trends [6],  [7]. It is specific to the battery materials composition and its fabrication process, which is imcompatible to other types of batteries [8].

Data-driven SOH estimation methods have gained recognition due to their excellent performance without looking into the battery internal mechanism. It takes readily measured variables as input and outputs the estimated capacities by finding the proper regression models and their associated parameters. In recent years, long short-term memory (LSTM), an artificial recurrent neural network created by Hochreiter $et\;al. $ [9], has been implemented into time series degradation prediction vastly. Zheng $et\;al. $ [10] presented a LSTM-based remaining useful life (RUL) prediction showing that LSTM notably outperformed traditional approaches like multi-layer perceptron, support vector regression, relevance vector regression. Jayasinghe $et\;al. $ [11] introduced a system model incorporating temporal convolutions with LSTM time dependencies to learn the salient features and complex temporal variations in sensor values to predict the RUL. Zhang $et\;al. $ [12] employed LSTM to capture the underlying long-term dependencies during the degradation and constructed a capacity-oriented RUL predictor.

Although LSTM is widely adopted by many researchers in degradation estimation,  it requires a standard length of input like other machine learning algorithms. There are several ways to scrub the data to have a uniform length to meet the system requirement like overall health indicator, manual truncation, zero padding,  etc. Specifically, Choi $et\;al. $ [13] constructed the input feature by taking partial samples from the entire cycling data. Liu $et\;al. $ [14] used health indicator by taking the time interval between the value of the variables or the value between the fixed time.  Manual truncation, which uses the shortest length as the reference, is widely adopted to cover as much data as possible.  Chen $et\;al. $ [15] investigated the impact of different window sizes on the prediction performance and concluded manual selection was needed based on different datasets for the LSTM attention mechanism.   Wu $et\;al.$ padded zeros to the header of the sequence to make the length of all sequences identical to the longest one [16]. 

However, compressing the whole cycle into samples or manual truncation involved human selection leading to information loss due to part of the information not being selected.  In LiB capacity prediction,  discharging time series data such as the voltage, current, temperature, etc.,  are commonly the input for machine learning algorithms.  In addition, each discharging cycle is not time-synchronized, which requires extra processing to make the machine learning approaches work. Although cycle synchronization is a matter regarding the accuracy of capacity prediction, few studies are available to deal with this important issue to the best of the authors' knowledge.

In order to solve the challenges mentioned above,  a cycle-synchronization-based SOH estimation model is being proposed in this paper.  It transforms the cycling data with uneven length in the original coordinate system to a new coordinate system by merging the time information into the original measurements. The benefits of this transformation are twofold.  Firstly, there is no physical information loss caused by manual operation or extension. Secondly, the synchronized time series enable the LSTM model to work smoothly without any extra data padding, which may cause performance reduction. Here, we use dynamic time warping (DTW) to achieve the above-mentioned approach and the final estimation model is developed by jointly combining the synchronized time series and LSTM.  Our main contributions are summarized as follows:

(1) The idea of cycle synchronization based on DTW is designed to deal with the uneven length cycling data of LiB for the first time, overcoming information loss due to manual selection of partial data.

(2) A SOH estimation framework is proposed to make the full use of battery degradation information, employing LSTM to capture the long time dependence.


(3) Experiments have been conducted to prove the efficacy of the proposed methods, further reducing the estimation error in comparision with the model without cycle synchronization.


The remaining of this article is organized as follows. Section II describes the capacity fading process and discharge cycle data. The proposed method is explained in Section III and followed by experiment results in Section IV. In the last section, the summary and the future work of this article are presented.

\section{Capacity fading process and data illustration}
During LiB usage,  the battery degrades along the execution of the charging and discharging process. The degradation rate is impacted by charge/discharge voltage, current, and surrounding temperature, which causes the rated capacity to decrease.  Fig. 1(a) describes a clear capacity fading trend of battery cell B18, which is provided by Ames Prognostics Center of Excellence [17]. The capacity has a decreasing trend along the discharge cycles, while capacity regeneration happens at certain cycles due to the rebalancing of active material inside the battery that causes capacity increase.

\begin{figure}[!htb]
\centering
\includegraphics[scale=0.31]{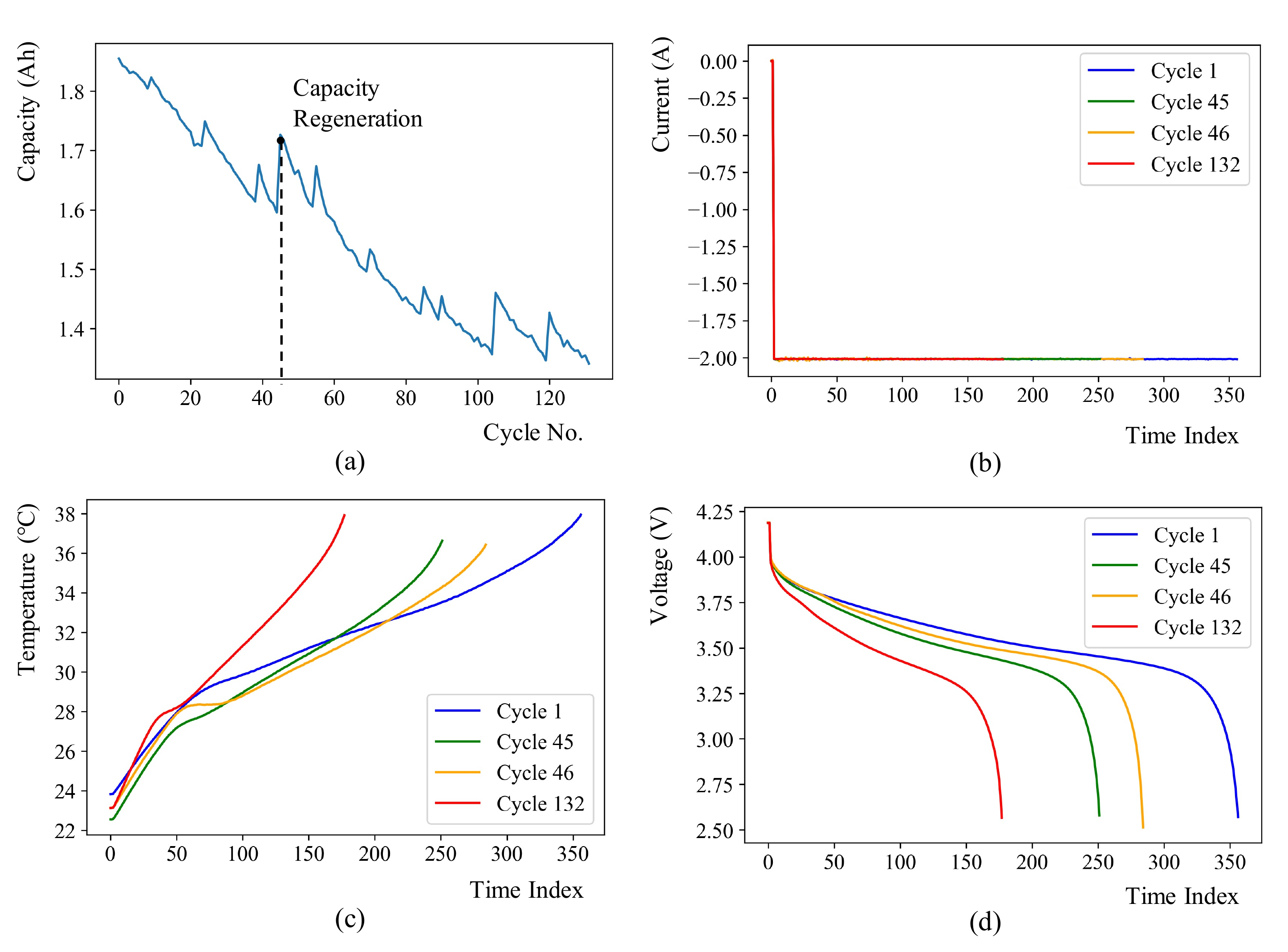}
\caption{Illustration of (a) Capacity loss over cycles, (b) discharge current signals, (c) temperature signals, and (d) voltage signals of the Battery 18 from the NASA dataset}
\label{MyFig1}
\end{figure}

Fig. 1(b), (c), and (d) illustrate the measured variables like current, temperature, and voltage discharge curves of Cycle 1, Cycle 45, Cycle 46, and the last cycle Cycle 132.  The time series of battery discharging data are presented under the coordinate system using time index as the X-axis and current,  temperature, voltage values as Y-axis at each discharge cycle.  Time index reduces as discharging occurs compared to Cycle 1.  Cycle 46 is the capacity regeneration cycle compared with Cycle 45. As a result of capacity surges in Cycle 46, the time index is longer compared with the one in Cycle 45. To sum up, time index is closely related to the battery's capacity level and is identified as battery degradation indicator.

\section{Methodology}
This section introduces the methodology of the proposed model with cycles synchronized for SOH estimation. Starting with existing coordinate system, the concept of coordinate system conversion is introduced. Next, the new coordinate system is formed through DTW by synchronizing the existing discharging cycles to be time-index-based time series. Finally, the synchronized time series are fed into LSTM to train the model for online SOH estimation. The overall flowchart is shown in Fig. 2.

\begin{figure}[htb]
\centering
\includegraphics[scale=0.45]{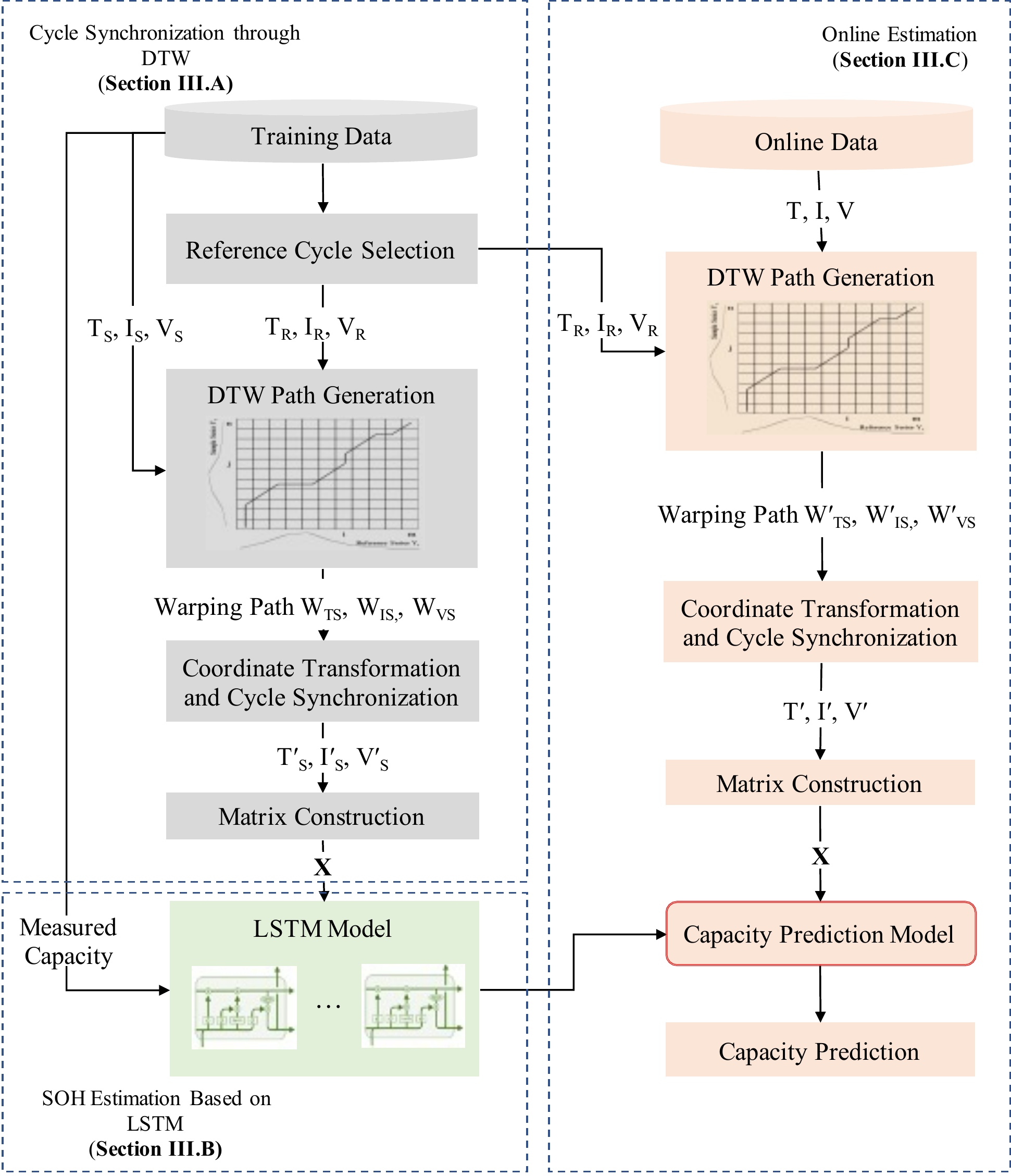}
\caption{Flowchart of battery performance degradation prediction by DTW coordinate transformation.}
\label{MyFig2}
\end{figure}

\subsection{Cycle Synchronization through DTW }
\subsubsection{Coordinate Transformation}
The coordinate system is used in geometry to identify the position of the point or other geometric elements on a manifold such as Euclidean space [18]. One of the widely used coordinate systems is the two-dimensional Cartesian orthogonal coordinate system.  It is defined by two perpendicular lines as the coordinate axis of X-axis and Y-axis with the intersect point as origin O.  To describe its position geometrically, point $\textbf p$ in the original coordinate system is denoted as $p\:(x,  y)$.  Under a different coordinate system,  the coordinate has changed to $p’(x’,  y’)$ after the original coordinate is transformed by the transformation matrix.  

Using time index to reflect battery degradation, a new coordinate system need to be developed by putting time index on Y-axis.  The X-axis in the new coordinate system is set up by the time index from the reference series \textit{R} so that all other series are having the same X-axis for standardization. The transformation relationship is shown in Fig. 3. Using voltage time series as illustration, the point in the original coordinate system $p(k_{s}^j,  V_{s}^j)$ from the sample series \textit{S} is converted to $p'(k_{r}^i,  k_{s}^j)$ using Eq. (1).  

\begin{equation}
\mathbf p'= 
\begin{bmatrix}
k_{r}^i \\
k_{s}^j
\end{bmatrix}
=\mathbf A\mathbf p
=\mathbf A\times
\begin{bmatrix}
k_{s}^j \\
V_{s}^j
\end{bmatrix}
\end{equation}
where $k_{r}^i$ is the time index from the reference series \textit{R},  $k_{s}^j$ is the time index from the sample series \textit{S},  $V_{s}^j$ is the voltage value under time index $k_{s}^j$ at the sample series \textit{S} and $\mathbf A$ is the transformation matrix.

In the new coordinate system, the reference cycle is converted to a diagonal line as the benchmark for other series and all the cycles are synchronized to have the same length as the reference series.

\subsubsection{Coordinate Transformation using DTW}
To get all the other series to have the same numbers of time index as the reference series,  all the time index $k_{r}^i$ in X-axis need to be filled by the time index from the sample series.  To find the respective $k_{s}^j$ to put in $k_{r}^i$,  it can be done by aligning the voltage value $V_{r}^i$ under $k_{r}^i$ to the closest value $V_{s}^j$ on the sample series and locate $k_{s}^j$ as illustrated in Fig. 3.
\begin{equation}
V_{r}^i\approx V_{s}^j
\end{equation}

\begin{figure}[htb]
\centering
\includegraphics[scale=0.32]{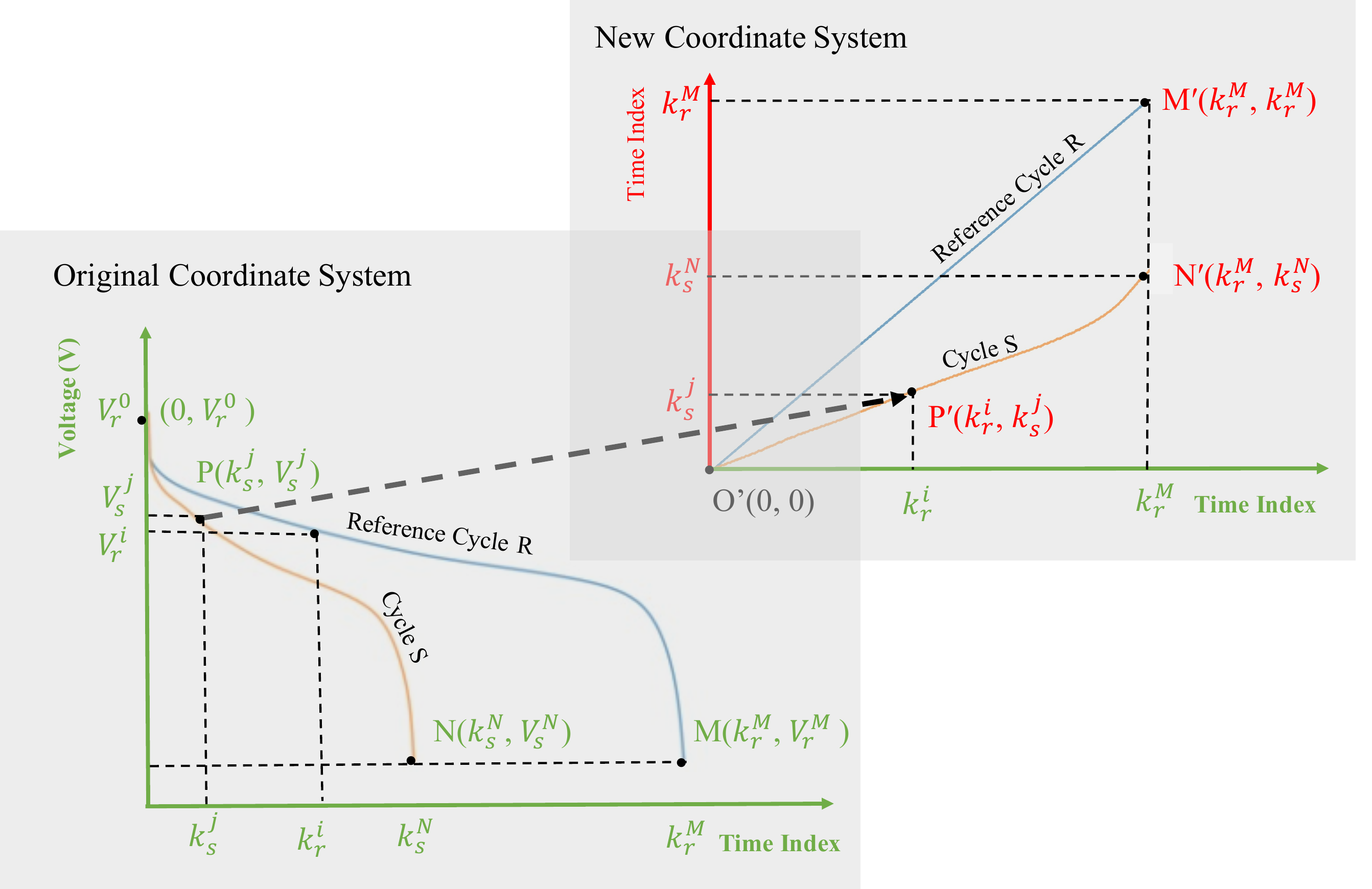}
\caption{Battery discharge cycles coordinate system transformation}
\label{MyFig3}
\end{figure}

DTW is well known for finding the optimal alignment between two time series. It has the ability to synchronize two trajectories by appropriately expanding and contracting localized segments to achieve a minimum distance between them [19].  As shown in Fig. 4(a),  a time index on sample series can be matched to multiple time indexes in reference series (green dashed line) and vice versa (red dashed line).

For a given reference series $\mathbf Y_R=[y_{r}^1,  y_{r}^2,...,y_{r}^M]$ and sample series $\mathbf Y_S=[y_{s}^1, y_{s}^2,...,y_{s}^N]$ ,  DTW constructs a path $\mathbf W$ based on minimum-distance between these two series.
\begin{equation}
\begin{aligned}
\mathbf W=[w_1, w_2, ..., w_3,...,w_Q]     
\end{aligned}
\end{equation}
where $Q$ is the length of the warping path and $\max(M,N) \le Q \le M+N$. 

The $q^{th}$ element of the waping path is 
\begin{equation}
\begin{aligned}
w_{q}=(k_{r}^i,k_{s}^j)    
\end{aligned}
\end{equation}

DTW maps the starting point and end point of the two series, where the path starts from $\mathbf w_1=(0,0)$ and end with $\mathbf w_Q=(M,N)$ .  Each time index of the time series must be used.  The path forming the synchronized relationship between the reference series and sample series is illustrated in Fig.  4(b).  

The following steps show the procedure of utilizing the DTW path to perform coordinate transformation for battery discharge cycle synchronization.

\textit{Step 1: Reference Cycle Selection}

A specific Cycle \textit{R} is picked as the reference cycle. Its time index $k_{r}^i$ is constructed as both X-axis and Y-axis. 

\begin{figure}[htb]
\centering
\includegraphics[scale=0.6]{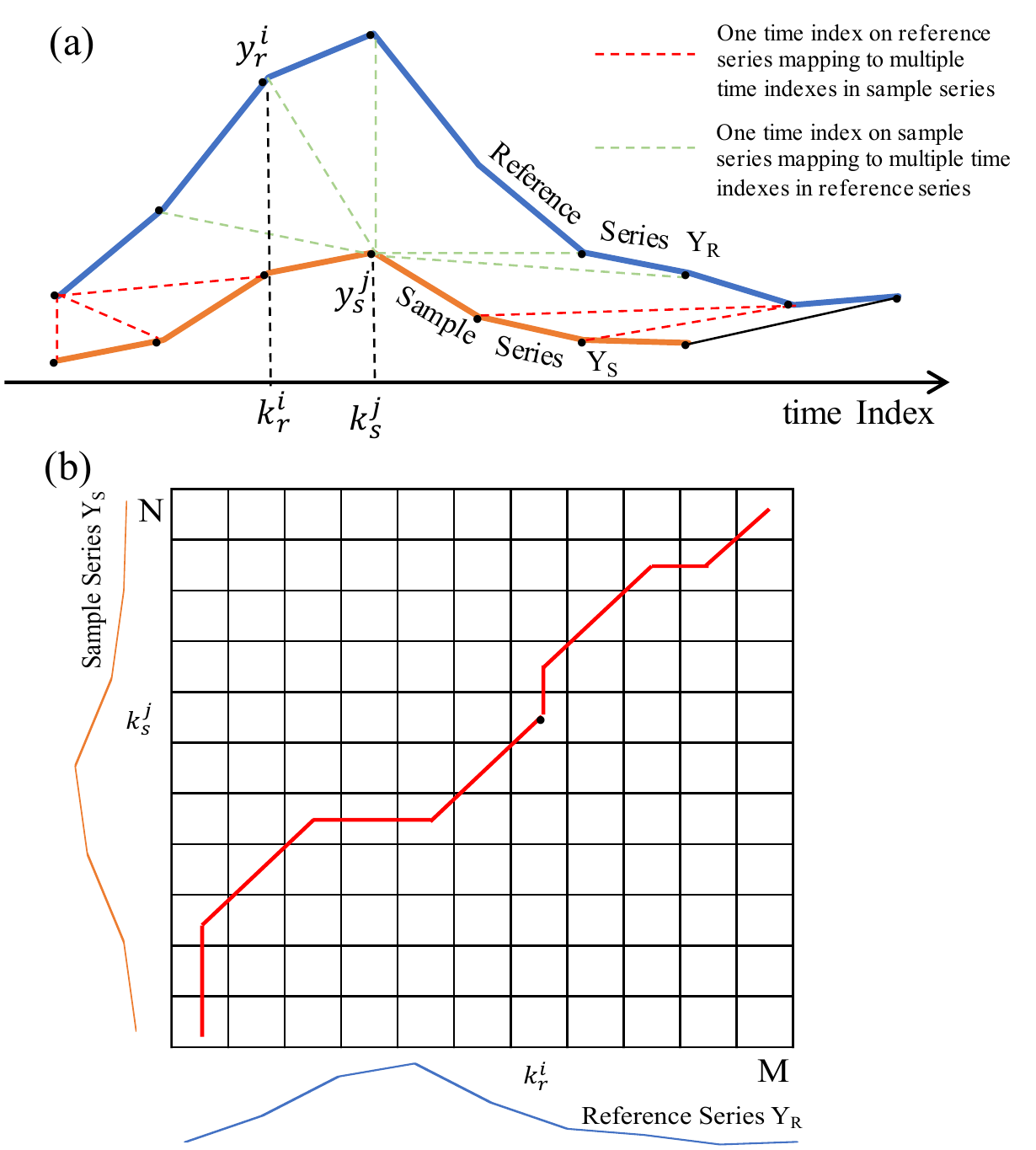}
\caption{Illustration of (a) Time series alignment by DTW and (b) DTW warping path}
\label{MyFig4}
\end{figure}
\textit{Step 2: DTW Path Generation}

DTW path is implemented here to find the time index $k_{s}^j$ in the samples series for the time index $k_{r}^i$.  As traditional DTW requires calculating $M\times N$ times to find the minimum distance [20],  FastDTW is adopted in this research to speed up the process.

Measured variables of temperature,  current, and voltage of the discharge cycles are denoted respectively as 
\begin{equation*}
\begin{aligned}
&\mathbf T_*=[T_{*}^1,  T_{*}^2, …, T_{*}^i, ..., T_{*}^j,… , T_{*}^K]\\
&\:\mathbf I_*\:= [I_{*}^1, I_{*}^2,…,I_{*}^i, ..., I_{*}^j,…,I_{*}^K]\\
&\mathbf V_*=[V_{*}^1, V_{*}^2,…,V_{*}^i, ..., V_{*}^j,…,V_{*}^K]
\end{aligned}
\end{equation*}
where the subscript * stands for either reference Cycle $R$ or a paired Cycle $S$, $\textit K$ refers to the length of the time series.

The warping paths between the temperature $\mathbf{T}_S$, current $\mathbf{I}_S$, voltage $\mathbf{V}_S$ of Cycle \textit{S} and their respective reference cycle $\mathbf{T}_R$, $\mathbf{I}_R$, $\mathbf{V}_R$,  are computed based on FastDTW. $\mathbf W_{TS}$,  $\mathbf W_{IS}$, $\mathbf W_{VS}$ are the generated DTW path for the temperature, current and voltage of Cycle \textit{S} accordingly. For each value of $w_{q}=(k_{r}^i,k_{s}^j)$, time index $k_{r}^i$ is put into X-axis and $k_{s}^j$  is put into Y-axis as illustrated in Fig. 5. 

\textit{Step 3: Cycle Synchronization}

Multiple time indexes synchronized to one time index is worked out here.  As demonstrated in Fig. 5, one time index $k_{s}^j$ in the sample series being assigned to multiple time indexes $k_{r}^i$ in the reference time series requires no further action. This is due to the reference time index $k_{r}^i$ is on the X-axis and its value is filled.  At the circumstance that multiple time indexes $k_{s}^j$ from the sample series are assigned to the same time index $k_{r}^i$ in the reference time series,  the mean of multiple time indexes ${\overline k}_{s}^j$ is used.

\begin{equation}
{\overline k}_{s}^j =\frac{\sum_{j=a}^{a+z}k_{s}^j}{z}
\end{equation}
where $\textit a$ refers to the starting point and $\textit z$ is the number of time indexes in sample series that are assigned to the same time index $k_{r}^i$ in reference series.

The illustration of the newly constructed time series is as following after all the $k_{r}^M$ time indexes are filled and all the time series are synchronized. 
\begin{equation*}
\begin{aligned}
&\mathbf T'_{S}=[k_{s}^1, ...\:, {\overline k}_{s}^j, {\overline k}_{s}^{j+1}, ...\:,\;k_{s}^N] \\
&\mathbf I'_{S}\;=[k_{s}^1,k_{s}^1,..., {\overline k}_{s}^j, ..., k_{s}^N,k_{s}^N] \\
&\mathbf V'_{S}=[k_{s}^1, k_{s}^2,k_{s}^2,..., {\overline k}_{s}^j, ..., k_{s}^N] 
\end{aligned}
\end{equation*}

\begin{figure}[htb]
\centering
\includegraphics[scale=0.4]{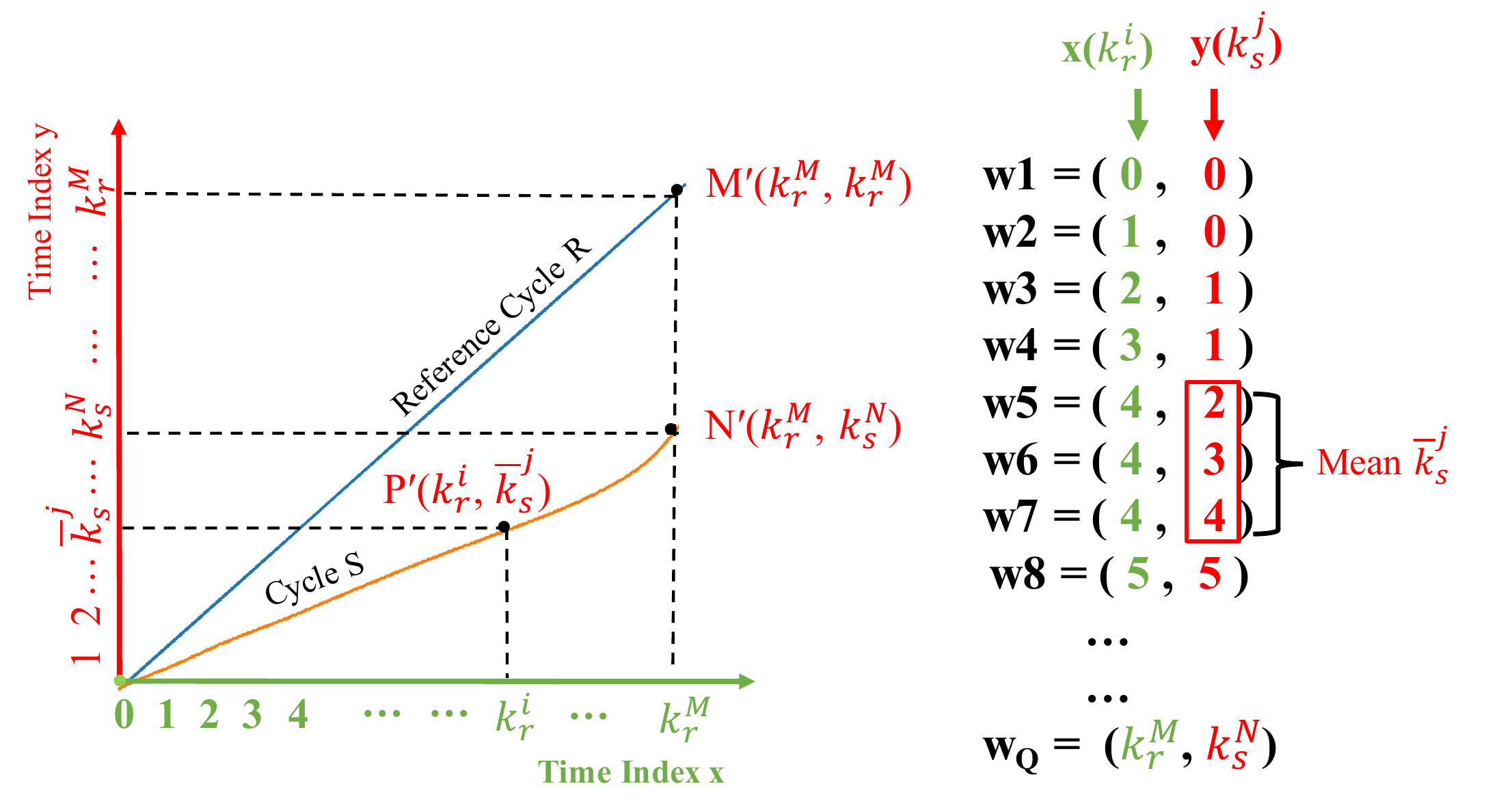}
\caption{New coordinate system by DTW warping path}
\label{MyFig5}
\end{figure}

\textit{Step 4: Matrix Construction}

The synchronized time series $\mathbf T'_{S}$,  $\mathbf I'_{S}$ and $\mathbf V'_{S}$ are constructed as $M$x$3$ input matrix for the machine learning prediction model as following.

\begin{equation}
\begin{aligned}
\mathbf{\mathbf X}
&= [\mathbf T'_{S},   \mathbf I'_{S},  \mathbf V'_{S}]
\end{aligned}
\end{equation}

\subsection{SOH Estimation Based on LSTM}
LSTM estimation model is selected to perform the prediction task.   It has internal mechanisms including cell states and gates to regulate the information flow. The cell state carries the information to maintain long-term memory of sequence, while the gates control what kind of information is allowed in the cell state.  The matrix generated from Eq. (6) will be the input to LSTM estimation model, where $\mathbf {x}_t$ is the $t^{th}$ sample of $\textbf X$. $\mathbf{h}_{t-1}$ is the hidden information from previous state. The first step is to decide what information to discard, which is controlled by the sigmoid layer named forget gate $\mathbf {f}_t$. The second step is to decide what information to store and it is done by $\mathbf{i}_t$ and $\mathbf {\tilde C}_t$.  Finally,  the new cell state $\mathbf {C}_t$ is formed. By combining $\mathbf {C}_t$ and the sigmoid function of input $\mathbf {x}_t$ and $\mathbf {h}_{t-1}$, it generates $\mathbf {h}_{t}$ as the hidden information for next state and as the output needed.  Detail algorithms are shown in Eq. (7).
\begin{equation}
\begin{aligned}
\mathbf {f}_t= \sigma(\mathbf {W}_f\mathbf {x}_t+\mathbf {U}_f\mathbf {h}_{t-1}+\mathbf {b}_f)\\
\mathbf {i}_t=\sigma(\mathbf {W}_i\mathbf {x}_t+\mathbf {U}_i\mathbf {h}_{t-1}+\mathbf {b}_i)\\
\mathbf {\tilde C}_t=tanh(\mathbf {W}_C\mathbf {x}_t+\mathbf {U}_C\mathbf {h}_{t-1}+\mathbf {b}_C)\\
\mathbf {C}_t=\mathbf {f_t}\times \mathbf {C_{t-1}}+\mathbf {i}_t\times \mathbf {\tilde C}_t\\
\mathbf {o}_t=\sigma(\mathbf {W}_o\mathbf {x}_t+\mathbf {U}_o\mathbf {h}_{t-1}+\mathbf {b}_o)\\
\mathbf {h}_t=\mathbf {o}_t\times tanh(\mathbf {C}_t)
\end{aligned}
\end{equation}

The predicted capacity $y_{pred}$ is coming from $\mathbf {h}_{t}$.  
\begin{align*}
         y_{pred} & = f(\mathbf {h}_t)
\end{align*}
where function $f(\cdot)$ is a combination of fully-connection layer and the ReLu operation.

The model is trained by minimizing the root mean squared error (\textit {RMSE}) between $y_{pred}$ and the actual capacity $y$ for all the inputs.
\begin{equation}
\begin{aligned}
\textit {RMSE}=\sqrt{\frac{1}{d}\Sigma_{i=1}^{d}{(y(i) -y_{pred}(i))^2}}
\end{aligned}
\end{equation}
where $d$ refers to the total number of the training cycles.

\subsection{Online Estimation}
After being trained by the offline data,  the model's parameters like weights, bias are well tuned to minimize the estimation error \textit {RMSE}.  When a new discharge cycle happens,  the time series of temperature $T$, current $I$, and voltage $V$ are fed into DTW separately to generate the path based on their respective first discharge cycle.  The new time-index-based time series $T'$,  $I'$,  $V'$ will be constructed based on the DTW path. Finally, they are put into matrix to feed into the well-trained LSTM model and output the predicted capacity.  More specifics of the online estimation could be found in Fig. 2.

\section{Experiment Result and Discussion}
\subsection{Cycle Synchronization by DTW}
In the dataset provided by NASA [17],  Battery18 was charged at a constant current of 1.5A until voltage raised to 4.2V, then continued in the constant voltage until charging current drops to 20mA. Discharge was carried out on a constant current of 2A until battery voltage drop to 2.5V from 4.2V.  It has 132 discharge cycles, and the first discharge cycle contains 357 time indexes.  Each interval between two time indexes was 10 seconds in real time.  With the discharging process occured,  the discharge cycle has been shortened along with the degradation that the last discharging cycle reduced to 178 time indexes.  

The first discharge cycle is chosen as the reference cycle where the battery is considered full capacity. Using the subsequent cycles to benchmark against it,  it could reflect the degradation status. The other discharge cycles, which have different time indexes, are transformed to the new series to have the same time index as Cycle 1 shown in Fig. 6. 

\begin{figure}[!htb]
\centering
\includegraphics[scale=0.45]{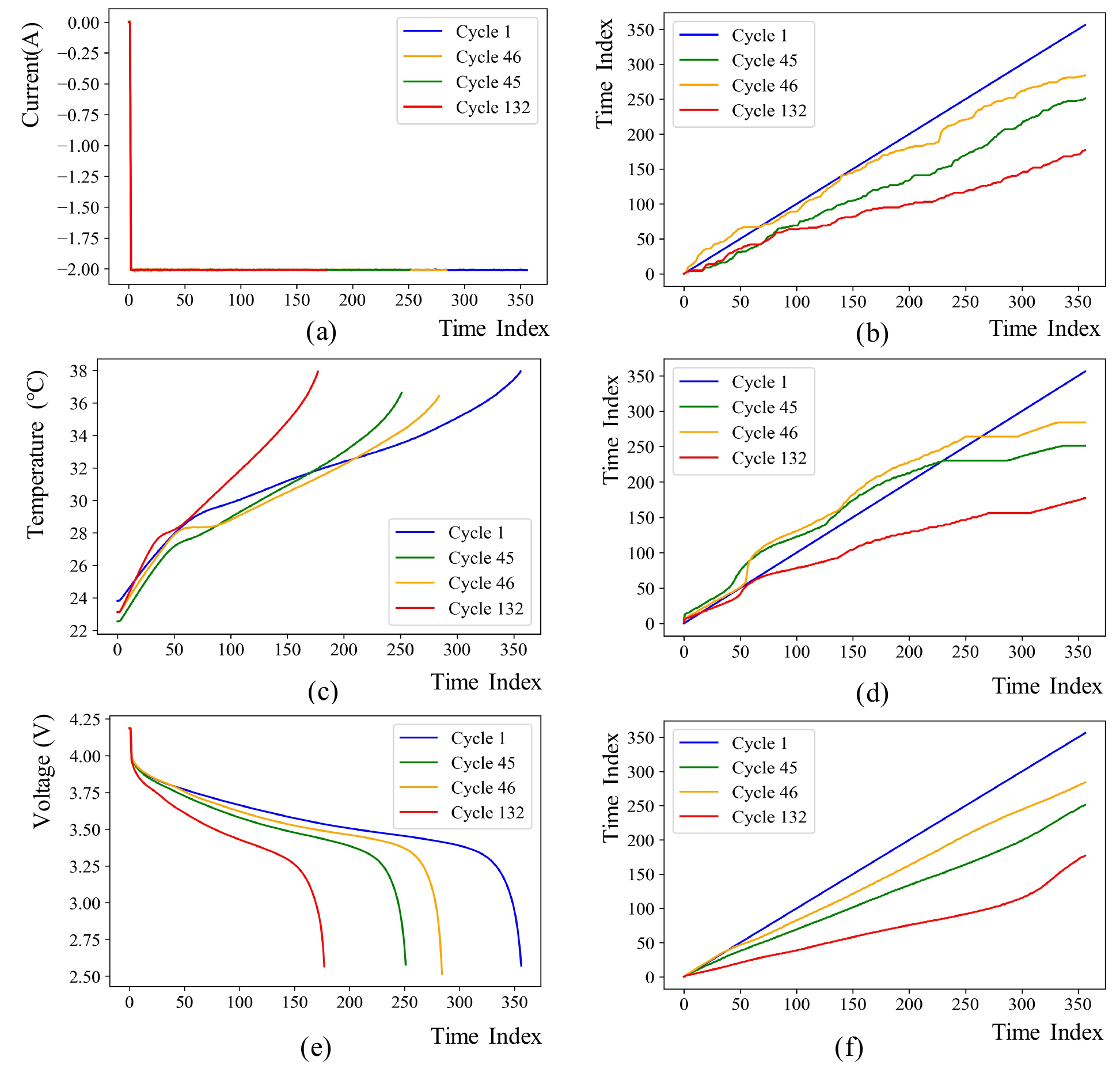}
\caption{Illustration of (a) Original discharge current signals,  (b) New discharge current series based on DTW path,  (c) Original discharge temperature signals,  (d) New discharge temperature series based on DTW path, (e) Original discharge voltage signals, and (f) New discharge voltage series based on DTW path.}
\label{MyFig6}
\end{figure}

\subsection{SOH Estimation Model using LSTM}
The LSTM estimation model is constructed by a series of LSTM layers and fully connected layers. Specifically, a LSTM layer of 200 neurons is followed by another LSTM layer of 300 neurons. And then, the aforementioned two LSTM layers are fully connected with a normal layer with 100 neurons before the output layer. The optimizer chosen here is Adam,  an algorithm based on adaptive estimates of lower-order moments for efficient stochastic optimization [21]. 100 epochs are run to find the optimum weights and biases.

\subsection{Performance Comparison with Cycle Synchronization}
To compare the proposed cycle-synchronization-based SOH estimation model, we manually truncate 178 time indexes counting backward from the original time series in this experiment. Typically, the total 132 cycles are separated into training data of 92 cycles and testing data of 40 cycles, which follows the ratio of 70$\%$ to 30$\%$ for training data to testing data. Moreover, to test the robustness of the proposed method, an estimation model is developed with a smaller amount of training data, which only employs the first 45$\%$ cycles at the training data.
\begin{table}[!htb]
	\scriptsize
	\renewcommand{\arraystretch}{1.5}
	\caption{comparison table for difference cases }
	\label{table_1}
	\begin{center}
		\begin{threeparttable}
			\begin{tabular}{c c c}
				\hline
				\toprule  
				{Name} & {70$\%$ Training} & {45$\%$ Training}\\
				\hline
				DTW-LSTM (Proposed) & 0.024 & 0.035 \\
				\hline
				Manual Truncation & 0.036& 0.056  \\
				\hline
				Performance Improvement \textit C & -33.3$\%$ & -37.5$\%$  \\
				\hline
				\toprule
			\end{tabular}
		\end{threeparttable}
	\end{center}
\end{table}

\subsubsection{Results with 70$\%$ Cycles as Training Data}
As shown in Table I, the estimation error of the proposed method is 0.024, which is much smaller than that of the manual-truncation-based LSTM model. For clarity, the proposed method is named DTW-LSTM here. The performance improvement between the two results is defined using the percentage changes formula as following.
\begin{equation}
\begin{aligned}
C = \frac{(\textit{RMSE}_{DTW}-\textit{RMSE}_{Manual})}{\textit{RMSE}_{Manual}} \times 100\%
\end{aligned}
\end{equation}
where $\textit {RMSE}_{DTW}$ indicates the proposed DTW-LSTM methods and $\textit {RMSE}_{Manual}$ is for the manual truncation methods. The negative outcome refers to \textit{RMSE} reduction, while positive outcome refers to \textit{RMSE} increment.

\begin{figure}[htb]
\centering
\includegraphics[scale=0.55]{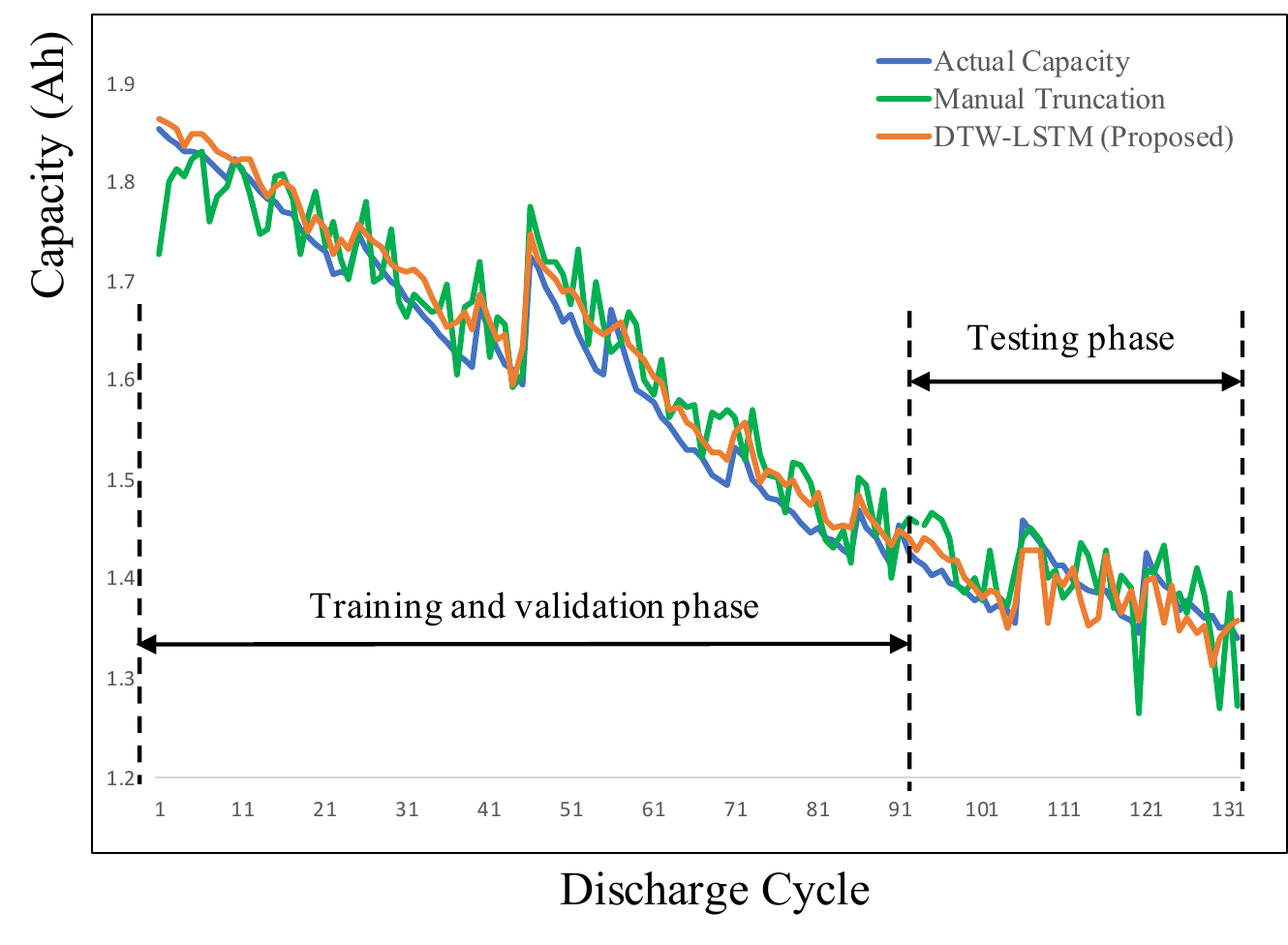}
\caption{Comparison between the proposed method with traditional LSTM with 70$\%$ training data}
\label{MyFig7}
\end{figure}

\begin{figure}[htb]
\centering
\includegraphics[scale=0.55]{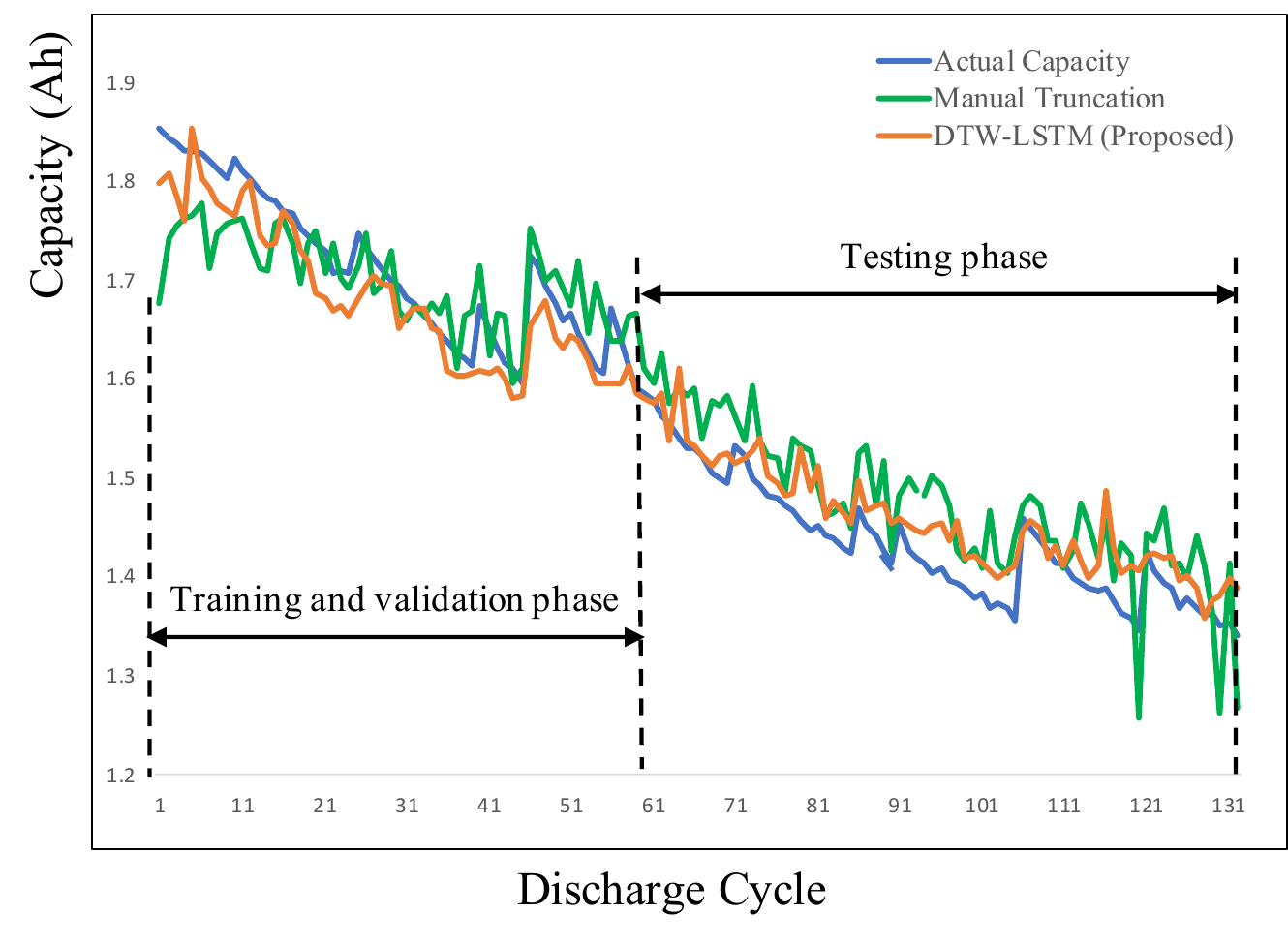}
\caption{Comparison between the proposed method with traditional LSTM with 45$\%$ training data}
\label{MyFig8}
\end{figure}

With the defined index $C$, it is observed that the proposed method could greatly reduce the estimation error by $33.3\%$, as listed in Table I.  Considering $RMSE$ is an overall index over cycles, the specific prediction results of the proposed method have further illustrated in Fig. 7, which are more smooth and close to the ground truth.

\subsubsection{Results with 45$\%$ cycles as training data}
By reducing the amount of training data to 60 cycles, the performance of the proposed method is further verified. In this case, DTW-LSTM performs better,  which \textit{RMSE} reduces by 37.5$\%$ as shown Table I. Also, the specific estimation results are provided in Fig. 8.

\section{Conclusion}
This paper presents an innovative insight into the importance of cycle synchronization to the time series in the LiB degradation process. It utilizes the complete information of the measured variables in cycles and turns them into time-index-based time series using DTW. With all the information being preserved, the experiment results prove that DTW-LSTM approach effectively identifies battery degradation progress compared with the traditional method.  Hence, it can be applied to other time series events which require identical data points to feed into the prediction system. Implementing this method into other machine learning algorithms would be the future of work to improve the battery degradation prediction further.

\vspace{12pt}
\end{document}